\documentclass[11pt, twocolumn, twoside]{IEEEtran}
\usepackage{cite}
\usepackage[cmex10]{amsmath}
\usepackage{fixltx2e}
\usepackage{datetime}
\usepackage{graphicx}
\usepackage{color}
\usepackage[T1]{fontenc}
\usepackage[utf8]{inputenc}
\usepackage{authblk}
\usepackage{amssymb}
\usepackage{algorithm}
\usepackage{rotating}
\usepackage{tikz}
\usepackage{enumerate}
\usepackage{tabularx}
\usepackage{verbatim}
\usepackage{pgf}
\usepackage{url}
\usepackage[margin=1.75cm]{geometry}
\usepackage{epstopdf}
\usepackage{soul}
\usepackage{algorithm}
\usepackage[noend]{algpseudocode}
\usepackage{amsmath}

\makeatletter

\newcommand{\Rmnum}[1]{\expandafter\@slowromancap\romannumeral #1@}
\makeatother

\makeatletter
\def\BState{\State\hskip-\ALG@thistlm}
\makeatother

\begin{document}
%
% paper title
% can use linebreaks \\ within to get better formatting as desired
\title{A deep-structured fully-connected random field model for structured inference}
%
%
% author names and IEEE memberships
% note positions of commas and nonbreaking spaces ( ~ ) LaTeX will not break
% a structure at a ~ so this keeps an author's name from being broken across
% two lines.
% use \thanks{} to gain access to the first footnote area
% a separate \thanks must be used for each paragraph as LaTeX2e's \thanks
% was not built to handle multiple paragraphs
%

\author{Alexander~Wong, \textit{IEEE Member}, Mohammad~Javad~Shafiee, Parthipan~Siva, and Xiao~Yu~Wang
\thanks{The authors are with the Vision and Image Processing Lab, Department
of Systems Design Engineering, University of Waterloo, 200 University Ave. West, Waterloo, Ontario, Canada, N2L 3G1. Tel.: +1 519 888 4567 x35342. Fax: +1 519 746 4791. E-mail: \{a28wong, mjshafiee, x18wang\}@uwaterloo.ca and psiva7@gmail.com.}}

% note the % following the last \IEEEmembership and also \thanks -
% these prevent an unwanted space from occurring between the last author name
% and the end of the author line. i.e., if you had this:
%
% \author{....lastname \thanks{...} \thanks{...} }
%                     ^------------^------------^----Do not want these spaces!
%
% a space would be appended to the last name and could cause every name on that
% line to be shifted left slightly. This is one of those "LaTeX things". For
% instance, "\textbf{A} \textbf{B}" will typeset as "A B" not "AB". To get
% "AB" then you have to do: "\textbf{A}\textbf{B}"
% \thanks is no different in this regard, so shield the last } of each \thanks
% that ends a line with a % and do not let a space in before the next \thanks.
% Spaces after \IEEEmembership other than the last one are OK (and needed) as
% you are supposed to have spaces between the names. For what it is worth,
% this is a minor point as most people would not even notice if the said evil
% space somehow managed to creep in.

% The paper headers
\markboth{IEEE Access journal}%
{A deep-structured fully-connected random field model for structured inference}
% The only time the second header will appear is for the odd numbered pages
% after the title page when using the twoside option.
%
% *** Note that you probably will NOT want to include the author's ***
% *** name in the headers of peer review papers.                   ***
% You can use \ifCLASSOPTIONpeerreview for conditional compilation here if
% you desire.

% If you want to put a publisher's ID mark on the page you can do it like
% this:
%\IEEEpubid{0000--0000/00\$00.00~\copyright~2007 IEEE}
% Remember, if you use this you must call \IEEEpubidadjcol in the second
% column for its text to clear the IEEEpubid mark.

% use for special paper notices
%\IEEEspecialpapernotice{(Invited Paper)}

% make the title area
\maketitle

\begin{abstract}
%\boldmath
There has been significant interest in the use of fully-connected graphical models and deep-structured graphical models for the purpose of structured inference.  However, fully-connected and deep-structured graphical models have been largely explored independently, leaving the unification of these two concepts ripe for exploration.  A fundamental challenge with unifying these two types of models is in dealing with computational complexity.  In this study, we investigate the feasibility of unifying fully-connected and deep-structured models in a computationally tractable manner for the purpose of structured inference.  To accomplish this, we introduce a deep-structured fully-connected random field (DFRF) model that integrates a series of intermediate sparse auto-encoding layers placed between state layers to significantly reduce computational complexity.  The problem of image segmentation was used to illustrate the feasibility of using the DFRF for structured inference in a computationally tractable manner.  Results in this study show that it is feasible to unify fully-connected and deep-structured models in a computationally tractable manner for solving structured inference problems such as image segmentation.
\end{abstract}
% IEEEtran.cls defaults to using nonbold math in the Abstract.
% This preserves the distinction between vectors and scalars. However,
% if the journal you are submitting to favors bold math in the abstract,
% then you can use LaTeX's standard command \boldmath at the very start
% of the abstract to achieve this. Many IEEE journals frown on math
% in the abstract anyway.

% Note that keywords are not normally used for peerreview papers.

% For peer review papers, you can put extra information on the cover
% page as needed:
% \ifCLASSOPTIONpeerreview
% \begin{center} \bfseries EDICS Category: 3-BBND \end{center}
% \fi
%
% For peerreview papers, this IEEEtran command inserts a page break and
% creates the second title. It will be ignored for other modes.
\IEEEpeerreviewmaketitle

\section{Introduction}

Structured inference, where the goal is to infer a structured state output from a structured observation input, is a crucial component for a wide range of applications such as object recognition~\cite{Quattoni}, image classification~\cite{Pan}, natural language processing~\cite{Yu1}, gesture recognition~\cite{Wang}, handwriting recognition~\cite{Feng}, and bioinformatics.  A powerful and commonly-used approach to structured inference is the use of Markov random field (MRF) and conditional random field (CRF)~\cite{Lafferty} models.  A limitation of such graphical models is that they utilize unary and pairwise potentials on local neighborhoods only, and as such can result in smoothed state boundaries as well as prohibit long-range state boundaries given the limitations of constraint locality.  This becomes particularly problematic in the presence of high observational uncertainties such as measurement noise and outliers.

Recently there has been significant interest in the application of two types of models for the purpose of structured inference that help address the issues associated with locally-connected graphical models: i) fully-connected graphical models, and ii) deep-structured graphical models.  Fully-connected graphical models address issues of locally-connected models by assuming full connectivity amongst all nodes in the graph, thus taking full advantage of long range relationships to improve inference accuracy.  One of the main hurdles in utilizing fully-connected graphical models is the complexity of inference, which becomes computationally intractable as the size of the problem scales.

Much of recent research in fully-connected graphical models have revolved around addressing the computational complexity of inference step.  Kr\"{a}henb\"{u}hl and Koltun~\cite{Koltun,Koltun2} introduced an efficient inference procedure for fully-connected CRF based on specific potential functions, where the edge potentials are obtained by use of Gaussian kernels, thus allowing them to formulate the inference problem as a filtering problem.  By computing the energy function via convolution, computational complexity is reduced to linear complexity by use of a permutohedral lattice~\cite{permu}.  Zhang and Chen~\cite{Zhang} utilized a stationary constraint where the spatial potentials over two nodes are assumed to depend only on their relative positions for each of their states, thus allowing for statistical encoding by different distributions and thus relaxing the Gaussian assumption made by Kr\"{a}henb\"{u}hl and Koltun.  Campbell et. al. \cite{Campbell} further generalized the pairwise potentials to non-linear dissimilarity measures by representing the pairwise terms as density estimates of the conditional probability. Ristovski et al. \cite{Ristovski} introduced a continuous fully-connected CRF that is similar to that proposed by Campbell et al., but targets the regression problems with continuous outputs.  Nevertheless, while the aforementioned methods significantly reduce the computational complexity of inference on fully-connected graphical models, they all address the problem similarly by defining specific potential functions to manage the inference as a filtering approach, thus limiting the effectiveness of such models as the key merit of such models is to allow for arbitrary feature function selection.

Deep-structured graphical models take a different approach to improving inference performance by introducing intermediate state layers, where there is a dependency of each layer on its previous layer, and inference is carried out in a layer-by-layer manner from bottom to top.  Prabhavalkar and Fosler-Lussier~\cite{Prabhavalkar} and Peng et al.~\cite{Peng} both introduced multi-layer conditional random field models where the local factors in linear-chain conditional random fields are replaced by multi-layer neural networks and trained via back-propagation.  Ratajczak et al.~\cite{Ratajczak} introduce a context-specific deep conditional random field model by replacing the local factors in linear-chain conditional random fields with sum-product networks.  Yu et al.~\cite{Yu1,Yu2} introduced a deep-structured conditional random field model which consists of multiple layers of simple CRFs where each layer's input consists of the previous layer's input and the resulting marginal probabilities.  While such deep-structured graphical models are good at handling high observational uncertainties such as measurement noise and outliers by characterizing different information at the different layers, they only implicitly take advantage of long range relationships and are more limited in this aspect when compared to fully-connected graphical models.

While fully-connected and deep-structured graphical models both have their own benefits and limitations, these two types of graphical models have been largely explored independently, leaving the unification of these two concepts ripe for exploration.  Such a unified graphical model could yield significant benefits in improving state boundary preservation, better enable long-range state boundaries, and better handle high observational uncertainties such as measurement noise and outliers.  A fundamental challenge with unifying these two types of graphical models is in dealing with computational complexity, as not only are all nodes fully-connected within a layer, there are also multiple layers to process due to the deep structure of the graphical model.  In this study, we investigate the feasibility of unifying fully-connected graphical models and deep-structured models in a computationally tractable manner for the purpose of statistical inference.  To accomplish this, we introduce a deep-structured fully-connected random field (DFRF) model which integrates a series of intermediate sparse auto-encoding layers placed between state layers to significantly reduce computational complexity while still maintaining the benefits of fully-connected and deep-structured graphical models.

This paper is organized as follows.  First, the methodology behind the proposed DFRF model and structured inference using DFRF for image segmentation is described in Section~\ref{methods}.  The experimental setup for evaluating the performance of the proposed DFRF model for solving the image segmentation problem is described in Section~\ref{setup}.  The experimental results and discussion is presented in Section~\ref{results}, and conclusions are drawn and future work discussed in Section~\ref{conclusions}.

%--------------------------------------------------------------------------------------------------------------------------------------------------
\section{Deep-structured fully-connected random fields}
\label{methods}
%--------------------------------------------------------------------------------------------------------------------------------------------------

From the Bayes rule~\cite{Bishop}, the joint distribution of the observation $X$ and labels $Y$ are modeled based on the product of conditional probability of labels given observation, $P(Y|X)$, and the probability of observation, $P(X)$ as
\begin{align}
P(Y,X) = P(Y|X) P(X).
\label{eq:joint-dist}
\end{align}

The goal of the proposed work is to incorporate fully connected interactions into the model that can preserve more information by taking advantage of the long range interactions in the modeling.  However, incorporating fully connected interactions imposes a high computational complexity into the model which makes inference intractable.  To address the issue of computational tractability, we introduce a sparse auto-encoding layer that describes the fully connected interactions among random variables more concisely with a smaller number of variables.  The auto-encoding layer is made possible as a result of the sparsity inherent in the structure of many types of data that are measured in a higher dimension than that needed to represent the data.  In essence the auto-encoding layer is representing the data as a smaller set of variables that describe the data in a more concise manner.

The interaction among variables are determined by extracting the interaction parameters from the auto-encoding layer variables instead of the variable $Y$. As a result Eq.~\eqref{eq:joint-dist} can be reformulated as
\begin{align}
P(Y,X) =  P(X) P(Y|X,A) P(X,A)
\label{eq:chain-rule}
\end{align}
\noindent where $A$ represents the auto-encoding layer where the number of its variables is much smaller than the number of output states. $P(X,A)$ characterizes the auto-encoding layer based on the observation and, based on the chain rule principle, is added to Eq.~\eqref{eq:chain-rule}. The role of the auto-encoding layer is to involve the fully connected relationship among nodes into the model implicitly.  The auto-encoding layer is constructed based on a specific number of variables, where the number of variables determines the fineness of structure in the data that can be characterized by the model (e.g., auto-encoding layers with fewer number of variables assume greater sparsity in the structure of the data and thus characterizes structure in the data more coarsely than when a higher number of variables are used).

\begin{figure}[tp]
	\centering
    \includegraphics[width=1\linewidth]{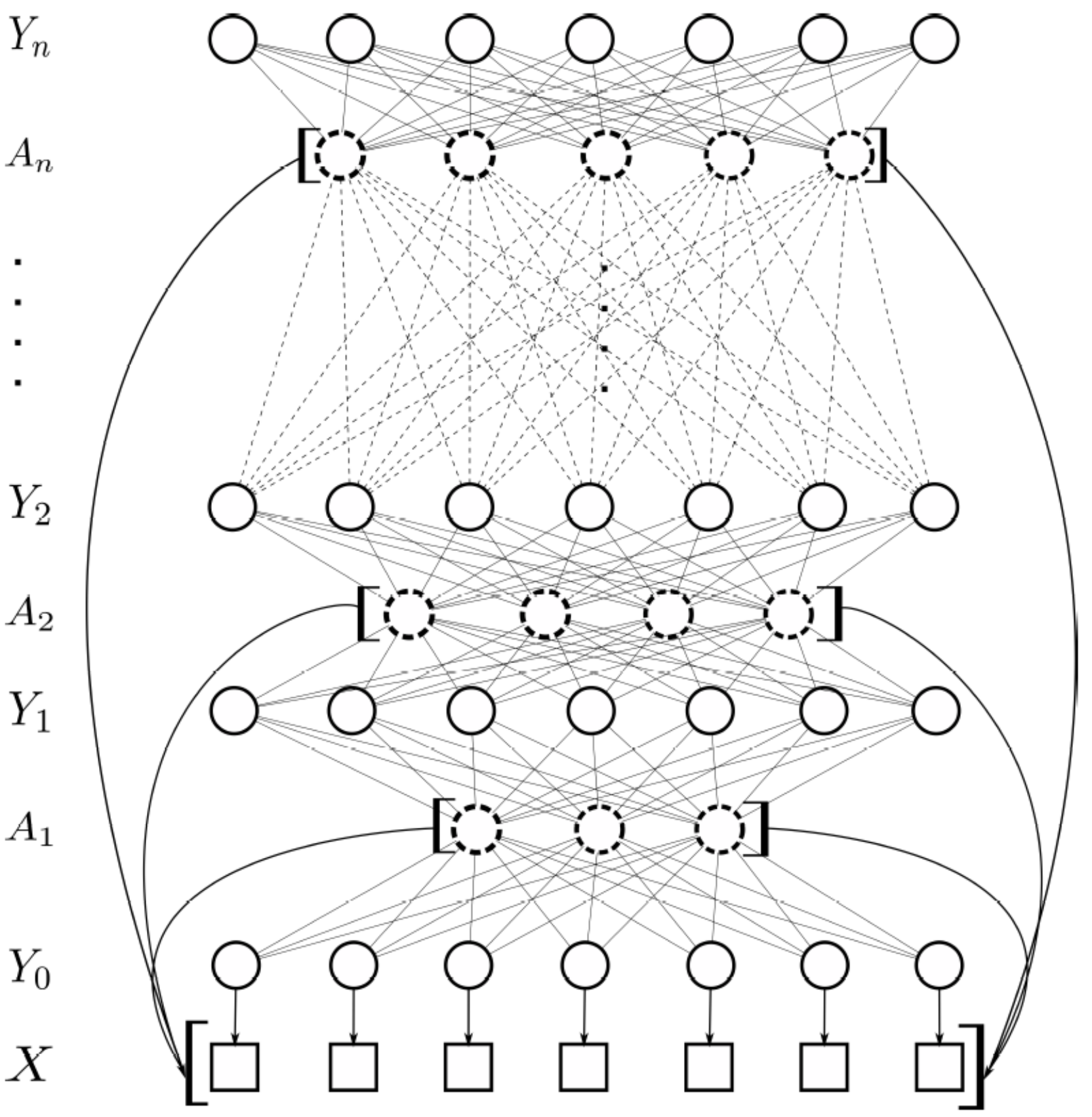}
	\caption{Deep-structured fully-connected random field model; The proposed framework is the combination of two different layers, auto-encoding layer ($A_i$) and label layer($Y_i$). The layer $Y_0$ is provided by finite mixture model (FMM) model and is an initialization for layer $Y_1$. Each node of layer $A_i$ is connected to all nodes in label layer $Y_i$. More information are provided to the model by increasing the number of nodes in the auto-encoding layers from the bottom to the top of the model. }
	\label{fig:deepModel}
\end{figure}

 To fully utilize the different concise structure characterization properties of the auto-encoding layer, a deep structure is used during the modeling. This results in the deep-structured fully-connected random field (DFRF) as shown in Fig.~\ref{fig:deepModel}. It is important to note that the configuration setup of the auto-encoding layers can be fine to coarse structure characterization or coarse to fine structure characterization depending on the specific application.  The coarse to fine configuration is utilized in this study for the problem of image segmentation described in the next section.

To represent the proposed model mathematically, the joint probability distribution of labels $Y$ and observation $X$ is formulated as a chain product of the conditional probability of labels given observation, auto-encoding variables, and previous layer of labels, multiplied by the joint probability of observation and auto-encoding variables (see Eq.~(4), where $Y_i$ represents the label layer of $i$, $A_i$ is the auto-encoding layer corresponding to layer $i$, and the number of layers is $n+1$).

\begin{figure*}[!t]
% ensure that we have normalsize text
\normalsize
\setcounter{equation}{3}
\begin{align}
P(Y,X) =  P(Y_n,X) = &P(X) \prod_{i=0}^n P(Y_i|X,A_i,Y_{i-1}) P(X,A_i)\\
 = &P(X) P(Y_0|X,A_0)P(X,A_0) \prod_{i=1}^n P(Y_i|X,A_i,Y_{i-1}) P(X,A_i) \nonumber\\
  = &P(X) P(Y_0|X) \prod_{i=1}^n P(Y_i|X,A_i,Y_{i-1}) P(X,A_i) \nonumber \\
 =  & P(X) \frac{P(Y_0,X)}{P(X)} \prod_{i=1}^n P(Y_i|X,A_i,Y_{i-1}) P(X,A_i) \nonumber\\
 =  & P(Y_0,X) \prod_{i=1}^n P(Y_i|X,A_i,Y_{i-1}) P(A_i,X) \nonumber
\end{align}
% Restore the current equation number.
\setcounter{equation}{4}
% IEEE uses as a separator
\hrulefill
% The spacer can be tweaked to stop underfull vboxes.
\vspace*{4pt}
\end{figure*}

\noindent Although there is no intra-layer connections among variables, the inter-layer interaction is fully connected and as such the interaction parameters among random variables in the label layer are computed by use of the auto-encoding layer. Therefore, the two aforementioned probabilities together are expressing a fully connected graphical model.

\subsection{Structured Inference using DFRF for Interactive Image Segmentation}

\begin{figure*}[tp]
\begin{center}
    \includegraphics[width=1\linewidth]{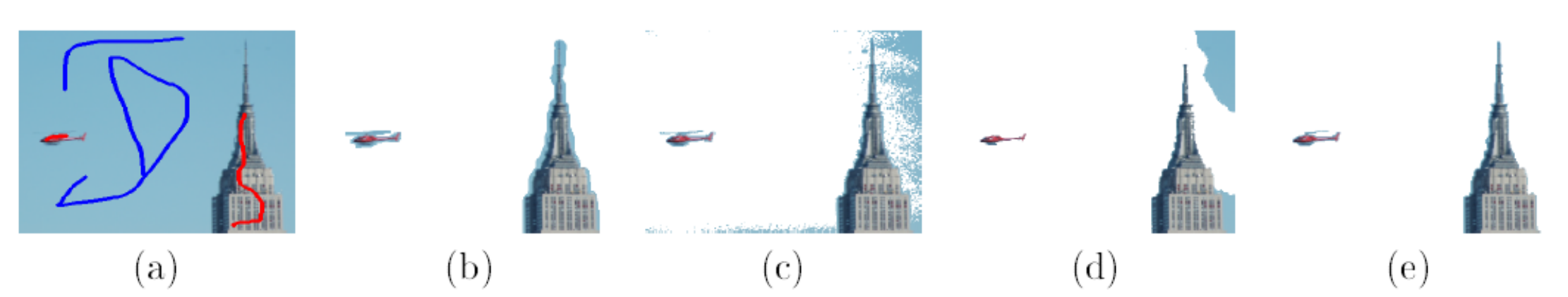}
\caption{Example of interactive image segmentation. \textbf{(a)} image with annotated seed regions (red: foreground and blue: background); \textbf{(b)} ground truth; \textbf{(c)} unary terms (GMM); \textbf{(d)} FCRF~\cite{Koltun}; and \textbf{(e)} DFRF. }
\label{fig:interactiveSeg}
\end{center}
\end{figure*}

We use DFRF for interactive image segmentation to illustrate the feasibility of DFRF for structured inference in a computationally tractable manner. Interactive image segmentation is a type of binary classification in which each pixel in an image must be classified as foreground (object) or background based on a small set of user annotated pixels as illustrated in Fig.~\ref{fig:interactiveSeg}.

A simple approach to tackle this problem is to learn a model based on the available training data, such as a Gaussian mixture model (GMM) or non-parametric histogram model, and apply the trained model to the image. However, this simple approach does not take into account the structure of the data. As a result, a common approach to tackle this problem is to use the trained model as the unary potential in a pairwise Markov random field (MRF) where the MRF enforces structural consistency.

Here, we utilize two finite mixture models (FMM) to model the background and foreground distributions and use them to define the first layer $Y_0$ in the deep structure model:
\begin{align}
P(Y_0,X) =  P(Y_0|\Lambda)  \text{\; \; s.t. \; \;} \Lambda = \{\mu_i ,  \sigma_i\}
\label{eq:FMM}
\end{align}
\noindent where $\Lambda$ is the set of trained mixture model parameters based on user annotated pixels. The results of layer $Y_0$ are propagated to the next layer (i.e. $Y_1$) by means of auto-encoding layer $A_1$. Each auto-encoding layer $A_i$ is constructed by maximizing the joint probability $P(A_i,X)$. The role of auto-encoding layer is to represent the structure of the image data in a concise manner using a smaller set of variables. Each auto-encoding layer characterizes the structural properties of the image data concisely at a certain fineness level as specified by the number of nodes in that layer. The joint probability $P(A_i,X)$ is modeled by a FMM as well:
\begin{align}
P(A_i,X) =  P(A_i|\Gamma)  \text{\; \; s.t. \; \;} \Gamma = \{\mu_i ,  \sigma_i\}
\label{eq:auto_encoding}
\end{align}
\noindent where $\Gamma$ represents the mixture model parameters. The number of parameters $|\Gamma|$ is  different for each auto-encoding layer based on the level the sparseness of the layer.

Each node in the auto-encoding layer conveys the interactions of a random variable in the label layer with other random variables based on a specific image data structure.  On the other hand, the interactions among random variables in a label layer $Y_i$ are expressed by the nodes in the lower adjacent auto-encoding layer $A_i$ that determines the weights and, therefore, the random variables in label layer $Y_i$ are fully connected implicitly.

The state of each random variable in the label layer $Y_i$ is obtained by a conditional probability given the auto-encoding layer $A_i$, observation $X$ and the previous label layer $Y_{i-1}$:
\begin{align}
P(Y_i|X,A_i,Y_{i-1}) =  \frac{1}{Z} \exp \Big(- E(Y_i|A_i,Y_{i-1}; X)\Big)
\label{eq:DFRF}
\end{align}
\noindent where the conditional probability is formulated as a Gibbs distribution~\cite{Gibbs} by the exponential of negative energy of the layer. $Z$ is the constant normalization and $E(\cdot)$ is the energy of the layer. The interaction weight between two random variables is computed based on their connections regarding to the auto-encoding layer.

Each random variable in the label layer $Y_i$ has two possible states, 0 and 1 determining the background or the foreground states. The energy in the layer $Y_i$ is minimized based on Maximum A Posterior (MAP) approach. The MAP framework tries to minimize the energy $E(\cdot)$ of layer $Y_i$ based on the observation and image data structural properties as characterized by auto-encoding layer $A_i$. The computed state configuration of layer $Y_i$ is passed to the layer $Y_{i+1}$ after each optimization. A step-by-step summary of image segmentation by DFRF is presented in Algorithm~\ref{Alg1}.

\begin{algorithm}
\caption{Structured Inference using DRFR for Image Segmentation}\label{euclid}
\begin{algorithmic}[1]
\Procedure{DFRF}{}
\State Set $n_{EV}$ \text{(The number of encoder variables)} and $n$ \text{(The number of layers excluding zeroth layer)}
\State $Y_0 \gets \arg\max P(Y_0|\Lambda)$
\State $i \gets 1$ ~~~~ (The layer number ($i$))
\BState \emph{loop}:
\State Find the auto-encoding layer $A_i($$n_{EV}$$),~Eq.~\eqref{eq:auto_encoding} $
\State $Y_i \gets \arg\max P(Y_i|X,A_i,Y_{i-1})~Eq.~\eqref{eq:DFRF} $
\State $i \gets i+1$
\State Increase $n_{EV}$
\State if i<=n then \textbf{goto} \emph{loop}
\BState \emph{endloop}
\EndProcedure
\end{algorithmic}
\label{Alg1}
\end{algorithm}

\section{Experimental setup}
\label{setup}

In this study, we use natural images to study the performance of the DFRF model for interactive image segmentation.  Natural images from the Weizmann Segmentation Evaluation Database~\cite{Weizmann} and the CSSD Complex Scene Database~\cite{CSSD} were used in this study.  The Weizmann database consists of two different datasets both with manual segmentation ground truth: i) a single-object dataset consisting of 100 images, and ii) a two-objects dataset consisting of 100 images.  The CSSD database consists of 200 images with manual ground truth.  Furthermore, to study binary classification performance at different noise levels, each of the images in the two datasets from the Weizmann database as well as the dataset from the CSSD database were also contaminated by white Gaussian noise with standard deviations at 25\% and 50\% of the dynamic range of the image, resulting in a total of 1200 different image permutations used in the analysis. A small set of seed pixels in the foreground and background are provided by the authors (Fig.~\ref{fig:interactiveSeg}(a)). All methods were compared based on the same annotated seed pixels.

To quantitatively evaluate segmentation performance, we compute the F$_1$-score as follows~\cite{Weizmann}:
\begin{equation}
f = \frac{2 \cdot TP}{2 \cdot TP + FN + FP}
\end{equation}
\noindent where $TP$ denotes true positive pixels, $FN$ denotes false negative pixels, and $FP$ denotes false positive pixels.

Our DFRF method is compared to the inference method for fully-connected CRFs proposed by ~\cite{Koltun} (which we will refer to as FCRF) using the implementation provided by the authors~\cite{Koltun}. FCRF is the state-of-the-art method for structured inference using fully-connected graphical models.  This method was also chosen for comparison because it had been shown~\cite{Koltun} that FCRF performs better than state-of-the-art approaches such as grid CRFs~\cite{Shotton} and $P^n$~CRF~\cite{pn-crf}. For a fair comparison, the same 5-component GMM model used for the DFRF is used as the unary potential for the FCRF approach.  %The results of directly using the 5-component GMM for the segmentation is also provided as a baseline solution which uses no structural information.

\subsection{Implementation details}

The DFRF has the following three parameters: i) the number of layers, ii) the number of encoding nodes at each sparse encoding layer (i.e., $n_{EV}$ (see Algorithm~\ref{Alg1})), and iii) the set of trained mixture models for layer $Y_0$ (i.e., $\Lambda$ (see Eq.~\eqref{eq:FMM})).  For our interactive image segmentation problem, we use 15 layers, and the number of encoding nodes at each sparse encoding layer is set to 450-660 nodes (increasing by $\sim$15 nodes between each layer and a 5-component Gaussian mixture model (GMM) is trained with the annotated samples and used for layer $Y_0$ in the DFRF. These parameters were found to provide strong classification performance based on comprehensive testing.

The DFRF is implemented in MATLAB (The MathWorks, Inc.) and can classify a 300 $\times$ 200 colour image (a total of 900,000 state nodes for this configuration) in $\sim$60s on an Intel(R) Core(TM) i5-3317U CPU at 1.70GHz CPU with 4GB RAM. The FCRF was implemented in C++ by the authors of~\cite{Koltun}, and can classify a 300 $\times$ 200 colour image in $\sim$0.48s.

\begin{figure*}[tp]
\begin{center}
    \includegraphics[width=1\textwidth]{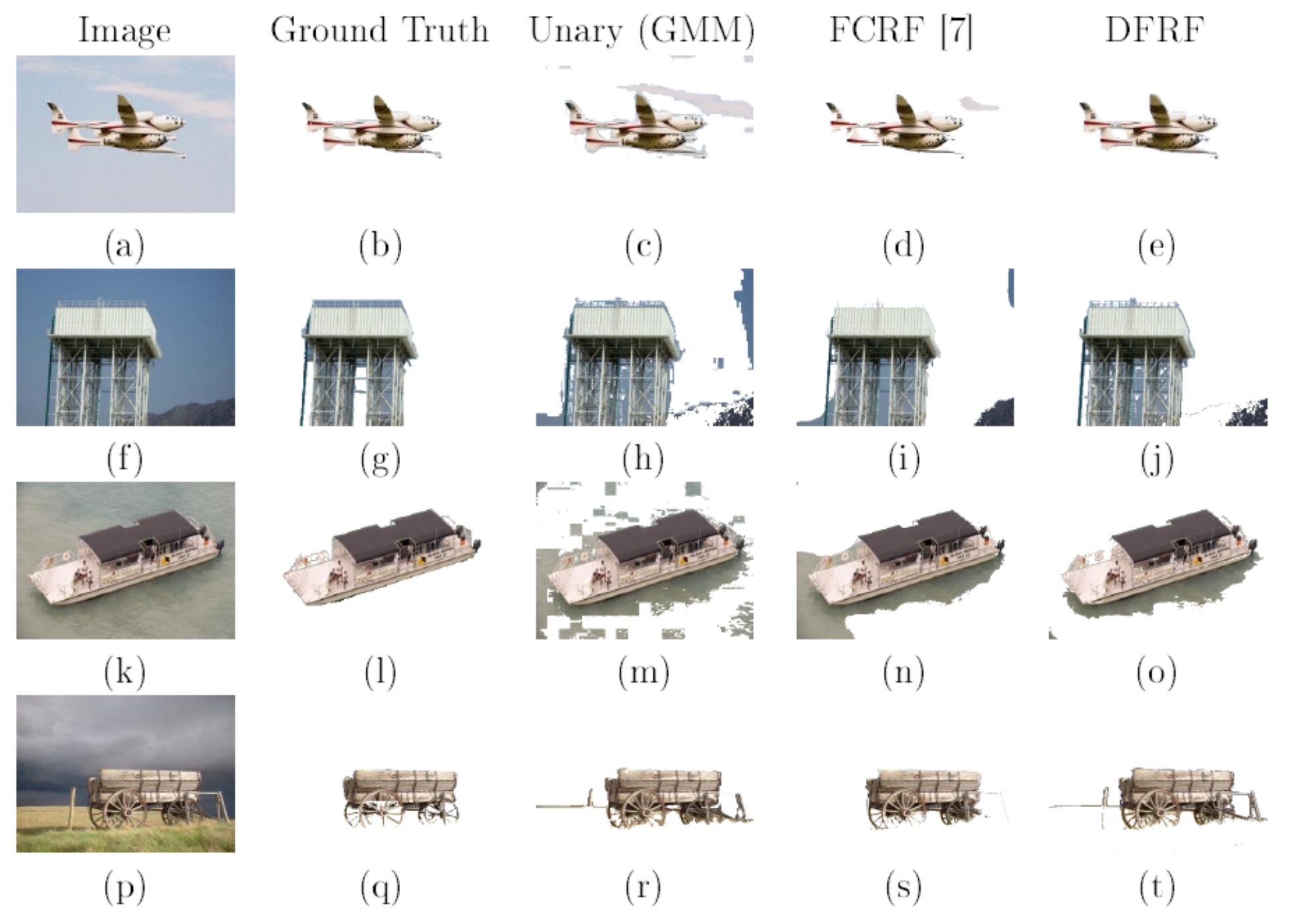}
\caption{Example segmentation results for single-object dataset. \textbf{(a,f,k,p)} image; \textbf{(b,g,l,q)} ground truth; \textbf{(c,h,m,r)} unary terms (GMM); \textbf{(d,i,n,s)} FCRF~\cite{Koltun}; and \textbf{(e,j,o,t)} DFRF.}
\label{fig3}
\end{center}
\end{figure*}

\section{Experimental Results}
\label{results}

The F$_1$-score achieved by the tested methods at the various noise levels for the Weizmann single-object dataset, the Weizmann two-objects dataset, and the CSSD dataset are shown in Table~\ref{tab:F1-scoreOneObj}, Table~\ref{tab:F1-scoreTwoObj}, and Table~\ref{tab:F1-scoreCSSD} respectively.  It can be observed that the binary image segmentation results produced using the DFRF model for the noise-free scenarios is comparable to the state-of-the-art FCRF method for the Weizmann single-object case. For the Weizmann two-object case, we see that DFRF performs slightly better than FCRF by $\sim$2\% for the noise-free scenario.  For the CSSD case, we see that FCRF performs slightly better than DFRF by $\sim$ 1\% for the noise-free scenario.

\begin{table}[tph]
	\centering
    \caption{F$_1$-Score for Weizmann single-object dataset.} % Classification performance results based on the Weizmann single-object dataset for noise-free scenario and two noisy scenarios.  The GMM (unary) result is provided as a reference.}
    \begin{tabular}{l||cccc}
    \hline
    ~ & 				   FCRF~\cite{Koltun} &	DFRF       \\ \hline
    Noise-free &        0.8655&   0.8606  \\
    Noisy (25\%) &        0.6586  &   0.6842\\
    Noisy (50\%) &        0.4959 &   0.5554 \\
    \hline
     \end{tabular}
    \label{tab:F1-scoreOneObj}
    \vspace{-0.3 cm}
\end{table}

\begin{table}[tph]
	\centering
    \caption{F$_1$-Score for Weizmann two-object dataset.} %Classification performance results based on the Weizmann two-objects dataset for noise-free scenario and two noisy scenarios.  The GMM (unary) result is provided as a reference.}
    \begin{tabular}{l||cccc}
    \hline
    ~ & 				   FCRF~\cite{Koltun}  &	DFRF      \\ \hline
    Noise-free &        0.8397  &   0.8594 \\
      Noisy (25\%) &        0.6718 &   0.7528 \\
    Noisy (50\%) &        0.5131 &   0.6030 \\
    \hline
     \end{tabular}
    \label{tab:F1-scoreTwoObj}
    \vspace{-0.3 cm}
\end{table}

\begin{table}[tph]
	\centering
    \caption{F$_1$-Score for CSSD dataset.} %Classification performance results based on the Weizmann two-objects dataset for noise-free scenario and two noisy scenarios.  The GMM (unary) result is provided as a reference.}
    \begin{tabular}{l||cccc}
    \hline
    ~ & 				FCRF~\cite{Koltun}  &	DFRF      \\ \hline
    Noise-free &    0.9558  &   0.9456 \\
      Noisy (25\%) &    0.8161 &   0.8462 \\
    Noisy (50\%) &    0.7122 &   0.7357 \\
    \hline
     \end{tabular}
    \label{tab:F1-scoreCSSD}
    \vspace{-0.3 cm}
\end{table}

\begin{figure*}[tp]
\begin{center}
    \includegraphics[width=1\textwidth]{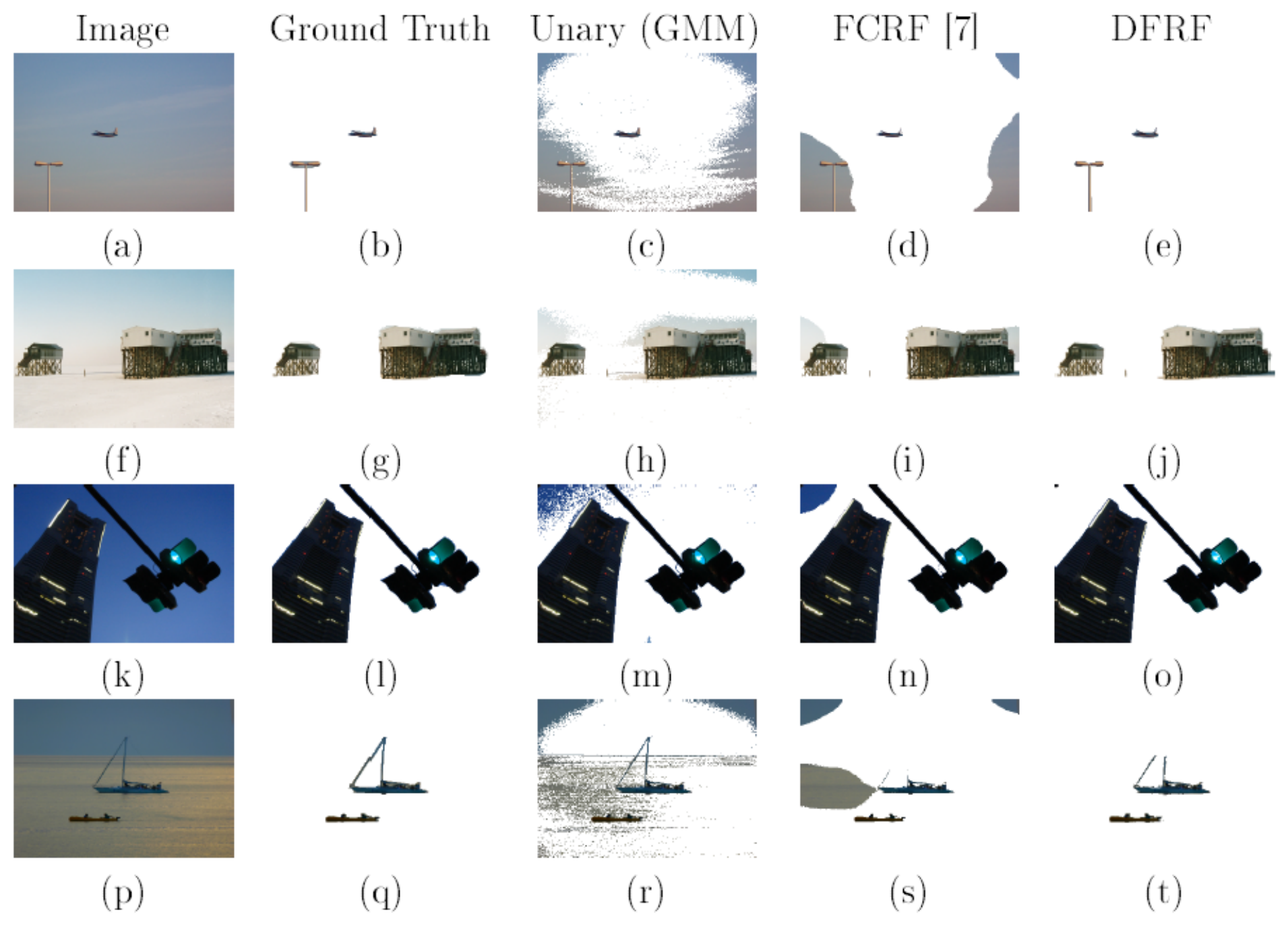}
\caption{Example segmentation results for two-objects dataset. \textbf{(a,f,k,p)} image; \textbf{(b,g,l,q)} ground truth; \textbf{(c,h,m,r)} unary terms (GMM); \textbf{(d,i,n,s)} FCRF~\cite{Koltun}; and \textbf{(e,j,o,t)} DFRF. }
\label{fig4}
\end{center}
\end{figure*}

Example segmentation results for Weizmann single-object and two-object datasets are shown in Fig.~\ref{fig3} and Fig.~\ref{fig4}, respectively. DFRF and FCRF preserve image structure much better than the baseline GMM method (used as unary) which has no structural cues. The lack of structural cues results in noise-like appearance in the segmentation as seen in Fig.~\ref{fig4}(c) and (r). Furthermore, the fully connected random field model allows both DFRF and FCRF to capture elongated and thin object boundaries (see the metal posts in Fig.~\ref{fig3}j, wooden fence post in Fig.~\ref{fig3}t, and the light post in Fig.~\ref{fig4}e).

Unlike FCRF, the deep-structure of our DFRF method allows us to handle slight variations in the observation that is not fully modeled by the small set of seeds provided by the user. For example, the sky and water in Fig.~\ref{fig4} have slight variation in illumination and texture which are not captured by the user annotation. As a result, the FCRF method starts misclassifying regions of the sky and water as foreground. However, our DFRF method is better able to handle these variations and correctly classify the entire sky and water as background.

\begin{figure*}
\begin{center}
    \includegraphics[width=1\textwidth]{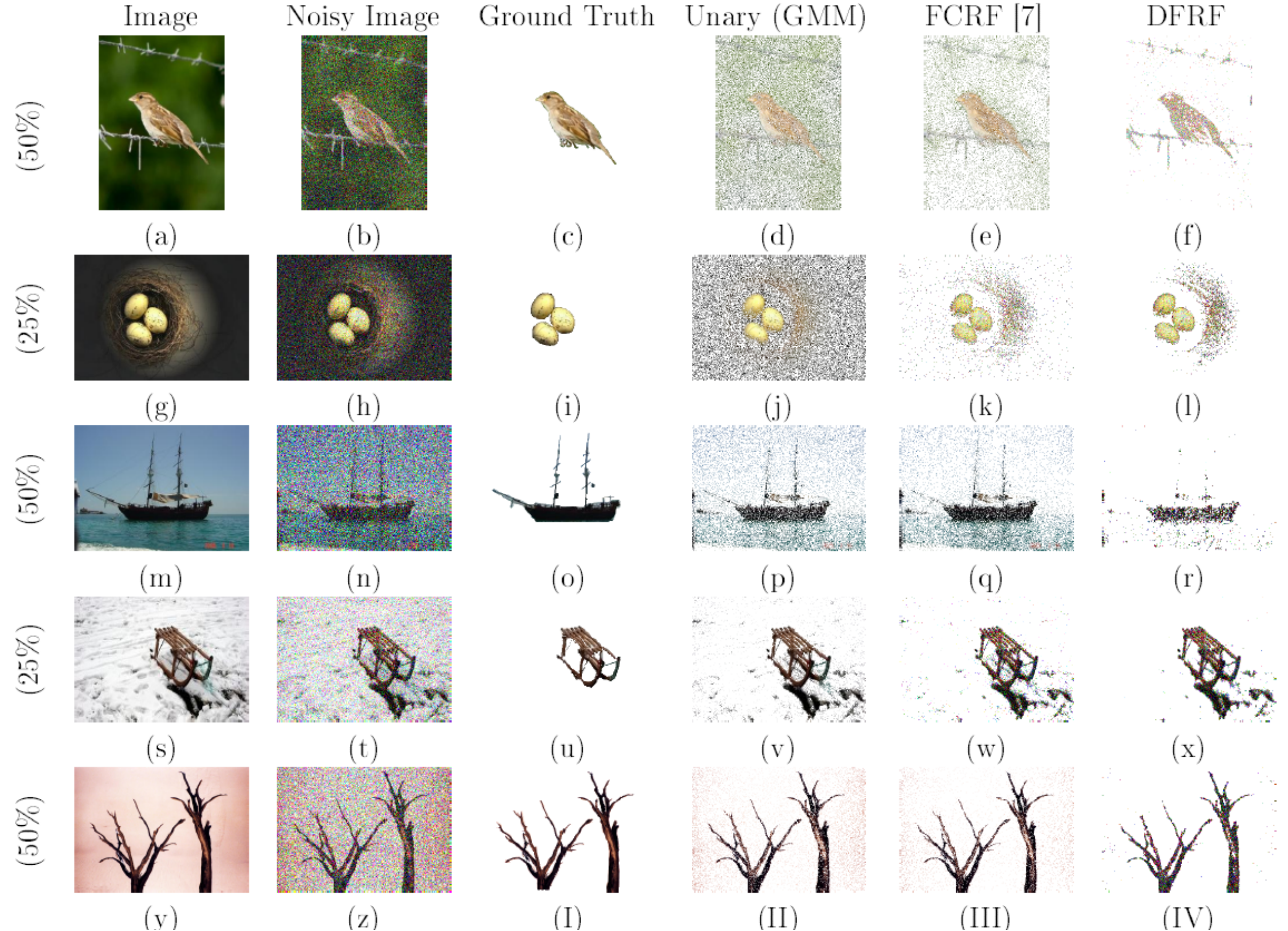}
\caption{Example segmentation results for noise-contaminated scenarios (noise level is indicated on the left side). \textbf{(a,g,m,s,y)} image;  \textbf{(b,h,n,t,z)} noisy image; \textbf{(c,i,o,u,I)} ground truth; \textbf{(d,j,p,v,II)} unary terms (GMM); \textbf{(e,k,q,w,III)} FCRF~\cite{Koltun}; and \textbf{(f,l,r,x,IV)} DFRF. }
\label{fig5}
\end{center}
\end{figure*}

DFRF's ability to handle variations in observation, due to its deep structure, can clearly be seen when we add noise to the image. Quantitatively, from Table~\ref{tab:F1-scoreOneObj}, Table~\ref{tab:F1-scoreTwoObj}, and Table~\ref{tab:F1-scoreCSSD}, we can see that DFRF clearly outperforms FCRF under the presence of noise for all datasets. Visually, from Fig.~\ref{fig5}, we can see that under the presence of noise FCRF starts to degrade and fails to maintain structural cues. On the other hand our DFRF method is able to handle the uncertainty in the observation and can better segment the image, even under presence of strong noise.

\section{Conclusion}
\label{conclusions}

In this study, the feasibility of unifying fully-connected and deep-structured models in a computationally tractable manner for the purpose of structured inference was investigated through the introduction of a deep-structured fully-connected random field (DFRF) model with sparse auto-encoding layers. By incorporating intermediate sparse auto-encoding layers between state layers to condense node-to-node interactions, we were able to significantly reduce the computational complexity of the inference process.  A quantitative performance analysis of the DFRF model for the problem of interactive image segmentation was performed to illustrate the feasibility of using the DFRF for structured inference in a computationally tractable manner.  Results in this study show that it is feasible to unify fully-connected and deep-structured models in a computationally tractable manner for solving structured inference problems such as image segmentation.

Given the promising results, we aim in the future to investigate alternative auto-encoding approaches to better condense node-to-node interactions, as well as strategies for automatically determining the number of auto-encoding nodes to use for each auto-encoding layer.  Furthermore, we aim in the future to explore the efficacy of the DFRF for solving other types of large-scale, vision-domain structured inference problems such as image reconstruction~\cite{recon1,recon2,recon3,recon4}, image decomposition and representation~\cite{imagerep1,imagerep2,imagerep3,imagerep4}, image restoration~\cite{restore1,restore2,restore3,restore4}, and saliency detection~\cite{saliency1,saliency2,saliency3}.

\section{Acknowledgment}

This work was supported by the Natural Sciences and Engineering Research Council of Canada, Canada Research Chairs Program, and the Ontario Ministry of Research and Innovation.

% Can use something like this to put references on a page
% by themselves when using endfloat and the captionsoff option.
\ifCLASSOPTIONcaptionsoff
  \newpage
\fi

\end{document}